\documentclass[sigplan]{acmart}

\usepackage[utf8]{inputenc}
\usepackage{amsmath, amsfonts}
\usepackage{enumitem}
\usepackage{array}
\usepackage{comment}
\usepackage{multirow}
\usepackage{subcaption}
\usepackage{makecell}
\usepackage{todonotes}
\usepackage[linesnumbered, ruled, vlined]{algorithm2e}
\usepackage{overpic}
\usepackage{booktabs}
\usepackage{colortbl}

\definecolor{codegreen}{RGB}{0,255,0}
\definecolor{codepurple}{RGB}{128,0,128}

\newcommand{\name}{NeuronMLP\xspace}

\definecolor{dong}{RGB}{0,0,255}
\definecolor{dinghong}{RGB}{30, 144, 255}
\definecolor{jierui}{RGB}{255, 165, 0}
\definecolor{pengfei}{RGB}{30, 144, 0}

\begin{document}



\title{\name: Efficient LLM Inference via Singular Value Decomposition Compression and Tiling on AWS Trainium}


\author{Dinghong Song}
\affiliation{%
  \institution{University of California, Merced}
  \country{USA}}  
\email{dsong15@ucmerced.edu}

\author{Jierui Xu}
\affiliation{%
  \institution{University of Wisconsin, Madison}
  \country{USA}}
\email{xjr2423@gmail.com}

\author{Weichu Yang}
\affiliation{%
  \institution{University of Wisconsin, Madison}
  \country{USA}}
\email{wyang338@wisc.edu}

\author{Pengfei Su}
\affiliation{%
  \institution{University of California, Merced}
  \country{USA}}
\email{psu9@ucmerced.edu}

\author{Dong Li}
\affiliation{%
  \institution{Yotta Labs \& University of California, Merced}
  \country{USA}}
\email{dli35@ucmerced.edu}

\begin{abstract} 

Emerging AI accelerators have started to gain attention and offer new opportunities for efficient inference of large language models (LLMs). 
Trainium, an AI accelerator recently developed by Amazon Web Services (AWS), provides an attractive option for LLM inference through its heterogeneous architecture. However, leveraging Trainium architecture for high performance can be challenging because of its systolic array architecture and special requirement on data layout. In this paper, we propose \name, an efficient LLM inference method based on Singular Value Decomposition (SVD) compression and tiling on AWS Trainium. We introduce a series of techniques customized to Trainium based on kernel fusion and novel caching strategies to reduce data movement across the software-managed memory hierarchy, maximize SRAM bandwidth, and avoid expensive matrix transpose. The proposed method is specifically optimized for multi-layer perceptron (MLP) layers in LLMs, which serve as a critical computational kernel for inference on Trainium. Evaluating on nine datasets and six recent LLMs, we show that \name significantly outperforms the state-of-the-art Neuron Kernel Interface (NKI)-based matrix multiplication (matmul) kernel implemented by AWS on Trainium: at the kernel level, it achieves an average 1.35$\times$ speedup, which translates to an average 1.21$\times$ speedup for end-to-end LLM inference, under a compression ratio of 0.05.

\end{abstract}%

\maketitle 

\section{Introduction}

Large Language Models (LLMs) have achieved remarkable success across a wide range of text-based tasks~\cite{liu2024deepseek,yang2025qwen3}.
Yet, the steady growth in their parameter counts and architectural complexity makes deployment increasingly prohibitive, particularly in resource-constrained environments. These challenges have driven extensive research into model compression~\cite{agarwal2024policy,frantar2023sparsegpt,dettmers2022gpt3,dettmers20218,sanh2020movement, zhang2020accelerating,li2023symbolic, chen2022disco} and efficient hardware design~\cite{wan2023efficient,zhu2024survey,jouppi2020domain,bai2023transformer,tzanos2022hardware,tuli2023acceltran,liu2023hardsea,ji2024beta,tuli2023acceltran}. 


AI accelerators~\cite{jouppi2023tpu, coburn2025meta, firoozshahian2023mtia}, customized to AI workloads, provide cost-effective and high-performance solutions for training and inference. Trainium is an AI accelerator recently developed by Amazon Web Services (AWS). It has been reported that Trainium can deliver 30–40\% lower cost while providing performance comparable to GPU-based EC2 instances~\cite{aws_trainium}. Each Trainium chip integrates two NeuronCores, each delivering up to 95 TFLOPS of FP16/BF16 compute capability, comparable to NVIDIA A100 GPUs at roughly 60\% of the cost. Such a cost–effective advantage makes Trainium an attractive platform for LLM training and inference~\cite{fu2024distributed, fan2024hlat}. Furthermore, Trainium, as a typical systolic-array architecture, features a programmable memory hierarchy, including two types of on-chip SRAM buffers, the State Buffer (SBUF) and the Partial Sum Buffer (PSUM), and an off-chip High-Bandwidth Memory (HBM). It also provides a rich set of specialized compute engines tailored for various AI operators. Such hardware heterogeneity gives programmers a lot of flexibility to explore for better performance.

However, leveraging Trainium architecture for high performance can be challenging. \textit{First}, as a systolic array architecture, Trainium must repeatedly go through a load-compute-store cycle to accommodate the small capacity of its on-chip SRAM. This design causes frequent data movement across the memory hierarchy, whose overhead can often be larger than the computation time spent in various compute engines in Trainium. In addition, allocating too much data to the on-chip SRAM can lead to implicit ``memory spill'' to HBM, which stalls the compute engines. On the other hand, underutilizing the on-chip SRAM wastes its high memory bandwidth and lowers the overall system throughput. \textit{Second}, the programmer must carefully align a tensor’s logical shape with Trainium’s physical memory layout.
Misalignment often requires tensor transposes and padding, which incur costly data transfers between HBM and on-chip SRAM.

In this work, we propose NeuronMLP, an efficient LLM inference solution based on Singular Value Decomposition (SVD) compression and tiling on AWS Trainium. By factoring a large weight matrix into two but retaining the top singular values and their corresponding singular vectors, we obtain a low-rank approximation to the original weight matrix. To make the application of SVD aligned with the capacity of on-chip SRAM in Trainium and maximize SRAM utilization, we employ a three-level hierarchical data layout (i.e, tile, block, and strip) and introduce a block-wise SVD. Applying SVD to the weight matrix $W$ (i.e., $W \approx UV$) decomposes the original large matmul ($XW$, where $X$ is input embedding) in LLMs into a sequence of two smaller matmuls, $XUV$. 
While this transformation reduces arithmetic complexity, a naive implementation on Trainium is bottlenecked by data movement and data layout transpose. Directly performing inference on SVD-compressed LLMs on Trainium leads to frequent data transfer and poor hardware utilization.
In particular, materializing the intermediate result (i.e., the output of the first matmul $XU$ in the sequence) exceeds on-chip SRAM capacity, forcing spills to HBM and subsequent reloads, which dominate execution cost. 
A natural optimization is to fuse the two matmuls to eliminate intermediate materialization. This approach recomputes required blocks of the intermediate result on demand, without storing them in HBM. 
However, on Trainium’s systolic-array architecture, this introduces a new I/O bottleneck: each recomputation requires reloading corresponding source blocks of $X$ and $U$, leading to redundant data movement that outweighs the benefit of avoiding materialization. Moreover, the layout mismatch between the two matmuls necessitates an intervening transpose, further increasing data movement overhead.

To address the above problem, \name introduces a new kernel fusion method for Trainium, named \textit{TrainiumFusion}. It is featured with three major techniques. First, TrainiumFusion introduces an SRAM-capacity-aware caching strategy to eliminate recomputation penalty. This strategy caches multiple rows of the intermediate matrix on the SRAM based on its capacity, and carefully reuses it when generating the output blocks with the corresponding column strips in the source blocks. This method avoids recomputation and frequent data movement. Second, TrainiumFusion reduces matrix transpose by leveraging the matrix-identify property without impacting the result correctness. Third, TrainiumFusion computes the matmul sequence by blocks in combination with its caching strategy and DMA-assisted result accumulation in SRAM. We further develop performance modeling to quantify the relationship between the block size and arithmetic intensity (or peak SRAM usage), allowing the programmer to maximize utilization of compute  engines while respecting the memory constraint.

We summarize the major contributions as follows.

\begin{itemize}

    \item We present \name, a high-performance matmul design for LLM inference on AWS Trainium that combines SVD-based compression with architecture-aware tiling. \name is specifically optimized for multi-layer perceptron (MLP) layers in Transformer, a dominant computational kernel in LLM inference.

    \item We introduce a series of Trainium-specific optimizations that minimize data movement across the software-managed memory hierarchy, maximize the utilization of SRAM and compute engines, and eliminate costly matrix transposes, and further improve hardware efficiency by applying tensor parallelism to SVD-decomposed weights. 

    \item We evaluate \name on nine datasets and six recent LLMs. Compared to the state-of-the-art AWS Neuron Kernel Interface (NKI)-based matmul kernel, at a compression ratio of 0.05, \name achieves an average 1.35$\times$  kernel-level speedup, translating to an average 1.21$\times$ end-to-end inference speedup with negligible accuracy loss.



\end{itemize}

\section{Background}



\subsection{SVD for LLM Compression}
\label{sec:bg}

SVD is a well-established technique for approximating high-dimensional matrices with low-rank representations~\cite{demmel1997applied}. Given a weight matrix $W$, SVD factorizes it into three matrices: $U$, $\Sigma$, and $V$, such that $W = U \Sigma V^\top$. By retaining only the top-$k$ singular values in $\Sigma$ and their corresponding singular vectors in $U$ and $V$, one obtains a low-rank approximation $W \approx U_k \Sigma_k V_k^\top$. This approximation preserves the most informative components of $W$ while substantially reducing the number of parameters to represent $W$. As a result, SVD is particularly well-suited for compressing the large and high-dimensional weight matrices in the MLP layers of LLMs, where parameter reduction directly improves efficiency with tolerable accuracy loss. 


Applying SVD involves two steps: matrix factorization and fine-tuning. We perform SVD offline on invariant weight matrices, and the resulting low-rank factors are used during inference for efficient matmuls. This design specifically targets LLM inference, where weights remain fixed and compression directly improves performance. 
In contrast, LLM training continuously updates weights, making repeated SVD factorization prohibitively expensive and thus impractical. As a result, \name is specifically optimized for inference rather than training.

\subsection{AWS Trainium}
AWS Trainium is a custom silicon chip designed to accelerate deep learning workloads. It adopts a systolic array–based architecture with rich hardware heterogeneity. Each Trainium chip integrates two NeuronCores. 
Each NeuronCore functions as an independent heterogeneous compute unit composed of 
a rich set of specialized engines designed for different operations, such as tensor engine, scalar engine, vector engine, and GPSIMD engine. 
In addition, Trainium provides direct memory access (DMA) engines to facilitate data movement between off-chip HBM and on-chip SRAM. These engines operate concurrently, enabling efficient execution across diverse deep learning workloads~\cite{aws_trainium_arch}. 
In this work, we focus on the tensor engine, which serves as the primary compute unit for matmul. 
We program Trainium using the Neuron Kernel Interface (NKI)~\cite{aws2024neuron}, a bare-metal language and compiler that enables direct control over Neuron devices. 

\textbf{Memory heterogeneity.} A NeuronCore is associated with three types of memory: 16 GB off-chip HBM, 24 MB on-chip State Buffer (SBUF), and 2 MB on-chip Partial Sum Buffer (PSUM). SBUF serves as the primary on-chip data buffer, while PSUM is a dedicated accumulation buffer for the tensor engine. Both SBUF and PSUM are organized as two-dimensional structures with 128 partitions. 
Computation proceeds by loading data from HBM into SBUF, where the data is accessible by all engines. Once the computation is completed, the final results are written back to HBM.  This explicit programming model shifts responsibility to software: efficient tiling and placement are essential to exploit on-chip locality. Inefficient management instead leads to excessive HBM traffic, prolonged DMA transfers, and tensor engine stalls, ultimately limiting performance on Trainium.
 
\textbf{Tensor engine.} It accelerates matmul by reading input tiles from  SBUF and writing output tiles to PSUM. Each tile-level matmul operates on two input matrices, referred to as the \emph{stationary} matrix and the \emph{moving} matrix (following AWS terminology). During execution, the stationary matrix is loaded into the tensor engine’s internal storage, while the moving matrix is streamed across it. Due to the systolic-array design, the stationary matrix must be consumed in a transposed layout~\cite{aws_trainium_arch}. 
We denote the low-level instruction \texttt{nki.isa.nc\_matmul} \cite{aws_nki_matrix_multiplication} as \texttt{NKIMatmul}, which carries out this tile-level operation. Accordingly, a matmul $AB$ is  formulated as \texttt{NKIMatmul(\allowbreak stationary = $A^T$, moving = $B$)}. 
The stationary matrix, which remains fixed during the computation, is transposed so that its rows align with the columns of the moving matrix, ensuring both input tiles share the same first dimension which corresponds to the partition dimension in SBUF. To formalize data organization, for a matrix $A \in \mathbb{R}^{M \times N}$, we define a three-level hierarchical layout, ordered from the finest to coarsest granularity: the Trainium-native \textit{tile}, the logical \textit{block}, and the matrix-spanning \textit{strip}.

\textbf{Matmul tiling.} Matrix multiplication on Trainium repeatedly performs the following three steps: (1) loading a pair of tiles from HBM into SBUF (stationary matrix must be transposed), (2) multipling the two tiles, and (3) storing the multiplication result into PSUM. PSUM is used to accumulate the tile-multiplication results from all input tiles. PSUM serves as a dedicated landing buffer for the tensor engine, with near-memory accumulation capabilities that enables read-accumulate-write operations at a fine granularity of every 4B memory elements. 
This accumulation mode is critical for large matmuls, particularly when the reduction dimension (i.e., the inner dimension in a matmul) is large.

\section{Motivation}

\subsection{Challenge 1: Data Movement}
\label{sec:data_movement}

\begin{figure}[!tbp]
  \centering
  \begin{minipage}[t]{\linewidth}
    \centering
    \includegraphics[width=1.0\linewidth]{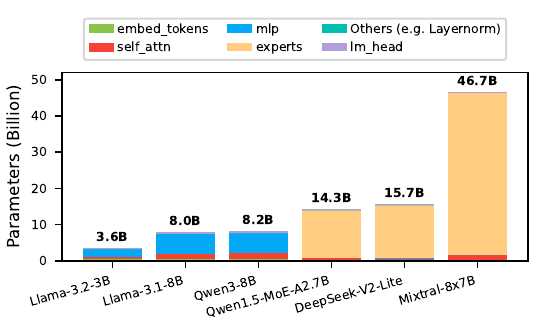}
    \caption{LLMs parameter breakdown by component. Whether for dense LLMs or MoE LLMs, the majority of model parameters are concentrated in the MLP layers or Expert layers. To simplify the comparison, we omit normalization modules (RMSNorm and LayerNorm) from the figure. 
    }
    \label{fig:model-paras-breakdown}
  \end{minipage}



\end{figure}

Figure \ref{fig:model-paras-breakdown} presents a parameter breakdown of different LLMs across their major components. For dense LLMs, the majority of parameters are concentrated in the MLP layers, accounting for as high as 70.2\% in Llama-3.1-8B, for instance. In MoE-based models, most parameters reside in the Expert layers, reaching up to 96.57\% in Mixtral-8x7B. Notably, both MLP and Expert layers primarily consist of three large and high-dimensional weight matrices, which dominate the overall parameter count. Each MLP layer or Expert module includes two paired weight matrices—the \texttt{up\_projection} and \texttt{down\_projection}—along with a \texttt{gate\_projection}. On AWS Trainium, matrix computations require preloading model weights and input hidden states from HBM into the on-chip SBUF. However, the limited capacity of SBUF leads to intermediate results exceeding on-chip memory when performing high-dimensional matrix multiplications, forcing spills to HBM. This incurs frequent data movement between on-chip SBUF and off-chip HBM during execution.

As established in Section~\ref{sec:bg}, SVD transforms a matmul $XW$ into a three-matrix chain $XUV$. However, executing this new formulation sequentially on Trainium introduces serious inefficiencies if not co-designed with the hardware architecture.


\begin{figure}[!tbp]
  \centering
  \includegraphics[width=\columnwidth]{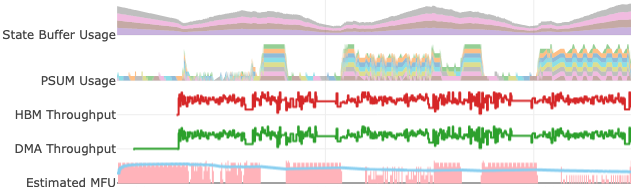}
  \makebox[\columnwidth]{Timeline} 
  \caption{The Neuron Profiler view of \texttt{up\_projection} matmul in Deepseek-V3 with SVD compression. Directly computing on the SVD-compressed weight matrices sequentially leads to low SBUF and PSUM utilization and reduced Model Float Utilization (MFU). Frequent idle periods in the MFU indicate that the tensor engine is underutilized while waiting for data transfers and data preparation to complete.}
  \label{fig:up_proj_svd_prof}
\end{figure}

Executing the SVD-compressed $XUV$ computation with a standard, sequential approach on Trainium results in poor hardware utilization. Figure~\ref{fig:up_proj_svd_prof} shows a Neuron Profiler trace of the SVD-compressed \texttt{up\_projection} matmul in Deepseek-V3~\cite{liu2024deepseek}. For an input length of 4096, with hidden size 7168 and intermediate size 18432, the \texttt{up\_projection} matrix ([7168, 18432]) is factorized into two low-rank matrices: $U$ ([7168, 4096]) and $V$ ([4096, 18432]). This transforms the original matmul ([4096, 7168] $\times$ [7168, 18432]) into a chain of three multiplications: [4096, 7168] $\times$ [7168, 4096] $\times$ [4096, 18432].
As highlighted in Figure~\ref{fig:up_proj_svd_prof}, this decomposition results in frequent idle gaps in Model FLOPs Utilization (MFU) and inefficient use of the on-chip SBUF, indicating that the tensor engine is often stalled waiting for data. 

The root cause is the materialization of the intermediate result. Sequential execution of $XUV$ first computes the intermediate matrix $Y = XU$, writes it from the fast on-chip SBUF to the slower HBM, and then reloads it into SBUF for the second multiplication $O = YV$. 
This 
\textit{load-compute-store} cycle creates a severe I/O bottleneck inherent to Trainium’s systolic-style architecture. 
Compared to the original \texttt{up\_projection}, the SVD-compressed version increases DMA transfer time by 65\% and more than doubles the traffic between HBM and SBUF, offsetting the benefits of reduced arithmetic. 

\begin{algorithm}[!tb]
\caption{Naive kernel fusion for $Y=XUV$.}
\label{alg:naive_fused_kernel}
\SetKwInput{KwInputs}{Inputs}
\SetKwInput{KwOutputs}{Outputs}
\SetKwInput{KwNote}{Note}
\SetKwFunction{FMatMulBlock}{MatMulBlock}
\SetKwProg{Fn}{Function}{:}{}

\footnotesize

\KwInputs{Matrices $X^T \in \mathbb{R}^{K \times M}$, $U \in \mathbb{R}^{K \times r}$, $V \in \mathbb{R}^{r \times N}$ in HBM.}
\KwOutputs{Result Matrix $O \in \mathbb{R}^{M \times N}$}
\Fn{\FMatMulBlock{$A, B$}}{
    \tcp{Calculate block matrix multiplication}
    \textbf{return} $A \times B$\;
}

Initialize output $O$ in HBM\;

\For{$m \leftarrow 1$ \KwTo $\lceil M/B_M \rceil$}{
    \For{$n \leftarrow 1$ \KwTo $\lceil N/B_N \rceil$}{
        $O_{mn} \leftarrow \mathbf{0}$\;
        \For{$p \leftarrow 1$ \KwTo $\lceil r/B_r \rceil$}{
            $Y_{mp} \leftarrow \mathbf{0}$\;
            \For{$k \leftarrow 1$ \KwTo $\lceil K/B_K \rceil$}{
                Load blocks $X^T_{mk}$ and $U_{kp}$ from HBM\;
                $Y_{mp} \leftarrow Y_{mp} + \FMatMulBlock(U_{kp}, X^T_{mk})$\;
            }
            $O_{mn} \leftarrow O_{mn} + \FMatMulBlock(Y_{mp}, V_{pn})$\;
        }
        Write $O_{mn}$ back to $O$ on HBM\;
    }
}
\textbf{return} $O$\;
\end{algorithm}

\subsection{Challenge 2: Recomputation}
To minimize data movement in computing $XUV$, we fuse the two matmuls into a single kernel, which we refer to as the \emph{naïve fused kernel}. 
Instead of materializing the intermediate result $Y=XU$ in HBM, the kernel recomputes $Y$ on the fly for each block of the final output. 
While this approach eliminates intermediate HBM writes, it shifts the bottleneck to increased computation and redundant HBM reads, as blocks of $X$ and $U$ must be repeatedly loaded to regenerate $Y$. 

Algorithm~\ref{alg:naive_fused_kernel} details our naive fused kernel.  The computational penalty arises because the calculation of the intermediate block $Y_{mp}$ (lines 9–11) is nested inside the main loop over $n$ (lines 5-13). As a result, the kernel redundantly recomputes the same $Y_{mp}$ for every output block $O_{mn}$ in a row. 
The computation of each output block $O_{mn}$ can be expressed as the following nested summation: 
\begin{equation}
\small
O_{mn} = \sum_{p=1}^{\text{NumBlocks}_r} \left( \underbrace{\left( \sum_{k=1}^{\text{NumBlocks}_K} X_{mk}U_{kp} \right)}_{\text{Intermediate block } (XU)_{mp}} V_{pn} \right)
\end{equation}

This formula reveals the source of recomputation. The inner summation, which calculates each block of the intermediate matrix $(XU)_{mp}$, depends only on the row-block index $m$ and the reduction-block index $p$; it is independent of the output column-block index $n$. Yet, the naive fused kernel re-calculates this inner summation for every output block in the row strip $O_{m,*}$ ($O_{m,1}, \dots, O_{m,\text{NumBlocks}_N}$). 
As a result, the same intermediate blocks are recomputed unnecessarily, incurring a penalty factor of $\text{NumBlocks}_N = \lceil N/B_N \rceil$.


This recomputation arises solely from Trainium’s on-chip memory limits. Avoiding recomputation would require setting $B_N = N$, producing an entire row-block $(B_M, N)$ at once.
But for LLM-scale matrices (e.g., $N = 16384$), such a block cannot fit within the 24 MB SBUF. In practice, we are therefore constrained to choose a much smaller $B_N$ (e.g., 512), forcing repeated recomputation. 
Worse, each recomputation triggers additional I/O: source blocks of $X$ and $U$ are repeatedly reloaded from HBM, inflating memory traffic and erasing the benefits of kernel fusion.

We quantify this penalty by benchmarking the naïve fused kernel against the sequential approach using matrix $X$ ([2048, 2048]), $U$ ([2048, 2048]), and $V$ ([(2048,8192]) with a block size of $B_N=512$. Table~\ref{tab:kernel_comparison} reports the results. The naïve fused kernel is more than 11$\times$ slower (18.06 ms vs. 1.57 ms). 
The slowdown stems directly from the recomputation factor $\lceil 8192/512 \rceil = 16$, which causes a 4$\times$ increase in FLOPs and a 10.5$\times$ increase in HBM traffic due to repeated reloading of $X$ and $U$.
This confirms that for LLM-scale matrices, recomputation penalties far outweigh the savings from avoiding intermediate I/O.

\begin{table}[!tbp]

\centering

\renewcommand{\arraystretch}{1.3}

\caption{Evaluation of the sequential matmul and naïve kernel fusion for matmul.}

\label{tab:kernel_comparison}

\resizebox{0.95\linewidth}{!}{%

\begin{tabular}{c|cc}

\toprule

\textbf{Metric} & \textbf{Sequential Matmul} & \textbf{Naive Kernel fusion} \\

\midrule

Total Time (ms)       & 1.57    & 18.06 \\

Model FLOPs (GFLOPs)  & 85.90 & 343.60 \\

{Memory Footprint (MB)} & 298.66 & 3140.42 \\

\bottomrule

\end{tabular}

}

\end{table}

\subsection{Challenge 3: Transpose Overhead} Transpose overhead on Trainium stems from the systolic-array design of its tensor engine, which requires the stationary matrix in a matmul to be supplied in a transposed layout. This requirement conflicts with the natural data flow of LLMs and introduces two types of overhead: (1) I/O transposes on tensors entering or exiting a kernel, and (2) intermediate transposes on on-chip temporary results. Our goal is to eliminate intermediate transposes entirely and minimize the cost of I/O transposes. 

\textbf{I/O transpose.} An I/O transpose arises when a tensor's layout required by an NKI compute kernel differs from the layout in which the tensor is stored in HBM, as dictated by the surrounding LLM computation graph. 
In the $XUV$ operator, for example, the input activation matrix $X$ must remain in its natural (non-transposed) layout to conform to the LLM data flow. However, for the first matmul $Y = XU$, the tensor engine requires the stationary input to be provided as $X^T$. Consequently, transposing $X$ is unavoidable, and thus constitutes an I/O transpose.

\textbf{Intermediate transpose.} An intermediate transpose arises when the output layout of one NKI kernel does not match the layout required by the next.
In the sequential execution of $XUV$, the first matmul produces $Y = XU$, which serves as the stationary input for the second multiplication $O = YV$. Because the tensor engine requires $Y$ in transposed form, a naïve implementation must explicitly materialize $Y^T$ on chip. 
This transformation incurs non-trivial overhead, as it introduces additional data movement within on-chip memory and synchronization costs.

\begin{figure*}[!tb] 
    \centering
    \begin{subfigure}{0.5\linewidth}
        \centering
        \includegraphics[width=\linewidth]{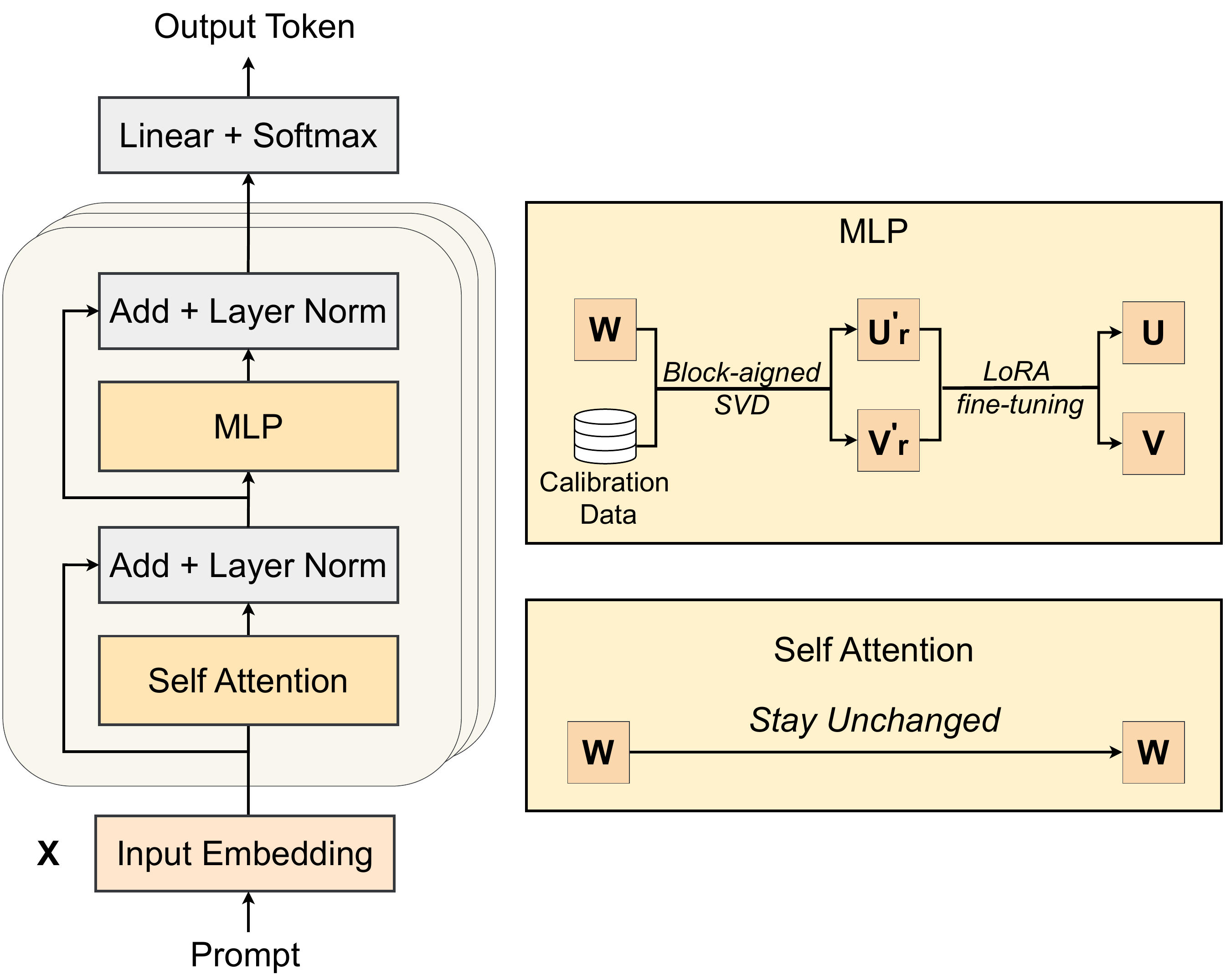}
        \caption{Block-aligned SVD}
        \label{fig:left}
    \end{subfigure}
    \hfill
    \begin{subfigure}{0.4\linewidth}
        \centering
        \includegraphics[width=\linewidth]{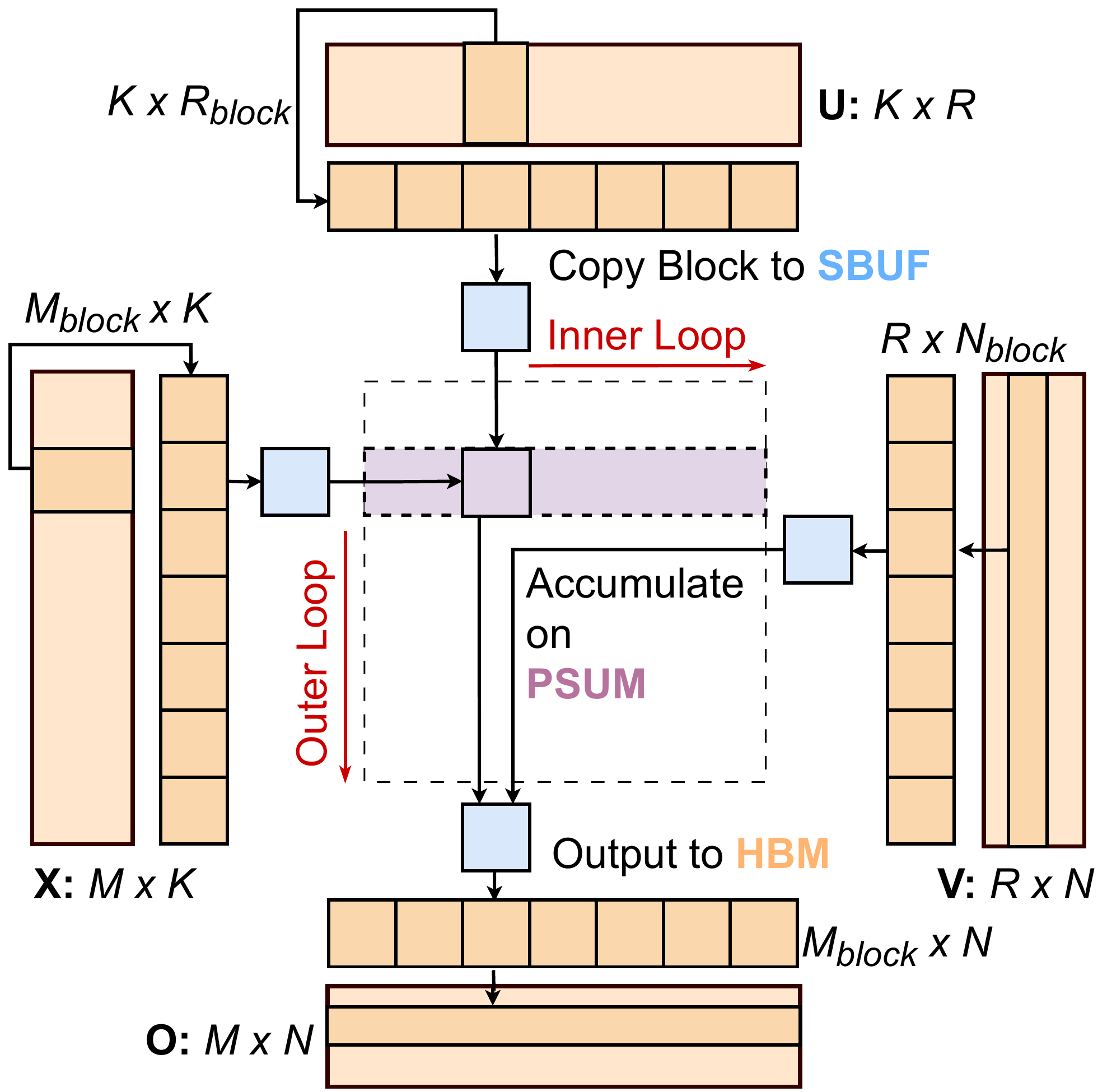}
        \caption{TrainiumFusion}
        \label{fig:right}
    \end{subfigure}
    \caption{The overview of \name.
\textbf{(a) Block-aligned SVD.} The weight parameters of the attention layers remain unchanged, while only the large  matrices $W$ in the MLP layers are compressed using SVD. \textbf{(b) TrainiumFusion.} 
The weight $W$ is decomposed into $U$ and $V$, and the original matmul $XW$ turns into $XUV$. The kernel leverages caching, implicit transposition, and blocking to enable efficient matmul, thereby reducing data movement between off-chip HBM and on-chip SRAM (SBUF and PSUM).
}

\label{fig:neuronmm_overview}
\end{figure*}


\section{Design}


Figure~\ref{fig:neuronmm_overview} provides an overview of \name, a framework for accelerating LLM inference on AWS Trainium. \name compresses the large weight matrices in the MLP layers using block-aligned SVD and restores accuracy through Low-Rank Adaptation (LoRA)~\cite{hu2022lora} fine-tuning. It further provides TrainiumFusion that execute matmuls on compressed models efficiently by exploiting Trainium’s architectural features.

\subsection{Block-Aligned SVD}
\textbf{Leveraging existing work.} Following the standard workflow of post-training LLM compression methods~\cite{yuan2023asvd,wang2024svd,ding2025dipsvd}, the weight matrix $W$ is first scaled by a matrix $S$ to capture the influence of input activations. $S$ is derived from a random set of input sentences. Specifically, for each MLP layer in LLMs, we record input activations $X$ using forward hooks, compute the covariance matrix, and apply Cholesky decomposition:
\begin{align}
SS^\top &= \text{Cholesky}(X^\top X)
\end{align}
where $S$ is a lower triangular matrix with all positive diagonal elements. \name then performs SVD on $WS$ and reconstructs $W$ using $S^{-1}$.
\begin{equation}
W = (WS )\cdot S ^{-1} = (U \Sigma V^\top) \cdot  S^{-1} \approx U_r \Sigma_r V_r^\top \cdot S^{-1}
\end{equation}
where $W \in \mathbb{R}^{k \times n}$, $U_r' \in \mathbb{R}^{k \times r}$, $V_r' \in \mathbb{R}^{r \times n}$, and $r$ denotes the top-r singular values. The choice of $r$ is crucial for balancing model accuracy and compression ratio. 

We define the \textit{compression ratio} as follows.

\begin{equation}
\text{ratio} = 1-\frac{\text{Model}\_\text{size}_{\text{compressed}}}{\text{Model}\_\text{size}_{\text{original}}}
\end{equation}
where $\text{Model}\_\text{size}_{\text{compressed}}$ and $\text{Model}\_\text{size}_{\text{original}}$ denote the model parameter counts with and without compression. This definition follows the existing work on LLM compression~\cite{wang2024svd,wang2025svd,li2025adasvd,yuan2023asvd,ding2025dipsvd}.  

\textbf{Block alignment.} Unlike prior work~\cite{wang2024svd, wang2025svd, wang2025dobi}, we introduce a block-aligned rank selection strategy that maximizes tensor engine utilization by coupling the compression ratio with the tensor engine’s $tile\_size$. The rank $r$ is computed as follows. 

\begin{equation}
r =  \left\lfloor \frac{k \times n \times (1-ratio) }{(k + n) \times block\_size } + \alpha \right\rfloor \times block\_size
\end{equation}
where $k$ and $n$ are the dimensions of $W$, and $block\_size$ is an integer multiple of $tile\_size$. \(\alpha\) is a rounding threshold that adjusts the required number of blocks, with $\alpha = 0.5$ corresponding to standard rounding. 
This formulation ensures that the choice of $r$ both satisfies the target compression ratio and aligns with hardware tile boundaries, avoiding intra-block padding and improving utilization of SBUF and the tensor engine. 
With $r$ determined, \hspace{3pt} $U_r,  \Sigma_r, V_r^\top$, and  $S^{-1}$ are consolidated into $U_r'$ and $V_r'$. 

\begin{align}
\footnotesize
U_r' &= U_r \cdot \sqrt{\Sigma_r} \\
V_r' &= \sqrt{\Sigma_r} \cdot V_r^\top \cdot S^{-1} \\
W &\approx U_r'V_r'
\end{align}
This transformation represents each weight matrix as the product of two low-rank matrices, reducing parameters from $kn$ to $r \times (k+n)$.

Although SVD substantially reduces the size of the weight matrix, it inevitably introduces accuracy loss. To recover accuracy, we apply LoRA fine-tuning to the compressed weights, similar to the prior work~\cite{wang2024svd,li2025adasvd}. During fine-tuning, we freeze the compressed  weights $U_r'$ and $V_r'$, and fine-tune them with LoRA:

\begin{align}
U \leftarrow U_r' + B_u A_u, \quad 
V \leftarrow V_r' + B_v A_v
\end{align}
where \(A_u, B_u, A_v\), and \(B_v\) are the trainable matrices used to adapt the model via LoRA. After fine-tuning, we incorporate the matrices $B_u A_u$ and $B_v A_v$ into $U_r^{\prime}$ and $V_r^{\prime}$, respectively, to form the final compressed weight matrices $U$ and $V$. 

\subsection{TrainiumFusion}
After SVD, we build a $XUV$ NKI kernel using new kernel fusion techniques in Trainium.




\subsubsection{XUV NKI Kernel}
\label{sec:xuv-kernel} 
As illustrated in Figure~\ref{fig:right}, we introduce three techniques: caching, implicit transposition, and blocking, to overcome the challenges of I/O bottlenecks and recomputation. The main idea is to fuse the $XUV$ chain into a two-stage pipeline that executes entirely within on-chip SBUF. 
In the first stage, we compute a strip of the intermediate product, using implicit transposition by reordering the inputs to the \texttt{NKIMatmul} primitive to directly produce its transpose, $(XU)^T$. 
This on-chip result is then immediately consumed in the second stage, where it is multiplied with a corresponding strip of $V$ to generate a block of the final output. 
This fused data flow eliminates intermediate data transfers between HBM and SBUF, avoids redundant recomputation, and removes the need for explicit intermediate transposes. We detail each technique below.

\textbf{Caching.} NeuronMM employs a capacity-aware caching strategy to eliminate recomputation penalty. The kernel calculates an entire row strip of the intermediate matrix, $(XU)_{m*}$, and caches it in a dedicated buffer within the on-chip SBUF. This cached strip is then efficiently reused for the subsequent multiplications with all corresponding column strips of the $V$ matrix ($V_{*1}, V_{*2}, \dots$) to produce every output block ($O_{m1}, O_{m2}, \dots$) to eliminate the recomputation. This on-chip caching is feasible within SVD-based LLMs, because the intermediate strip's memory usage is manageable. In particular, its shape is $(B_M, r)$, where $B_M$ is a block size (e.g., $1024$) and $r$ is the SVD rank. The resulting buffer size (e.g., $1024 \times r \times 2$ bytes for \texttt{float16} tensor) fits within the 24 MB SBUF, leaving sufficient capacity for other working data.

\textbf{Implicit transposition.} The low-level \texttt{NKIMatmul} primitive computes a matmul $AB$ using a constrained input layout, \texttt{NKIMatmul(stationary=$A^T$,} \texttt{moving=$B$)}, as required by the tensor engine. 
In a naïve implementation of $O=(XU)V$, the intermediate result $Y = XU$ is first computed, and the subsequent multiplication $O=YV$ requires $Y^T$ as the stationary input.
This layout mismatch necessitates an explicit transpose of $Y$, introducing overhead that scales linearly with the input sequence length. 
\name eliminates this overhead via implicit transposition. 
By leveraging the matrix identity $(XU)^T = U^T X^T$, we directly compute the transposed intermediate using \texttt{NKIMatmul(}\texttt{stationary=$U$,} \texttt{moving=$X^T$)}. 
In practice, this is realized by controlling the order of the inputs to the \texttt{NKIMatmul} primitive, producing $Y^T$ on chip without any explicit data transpose. 


\textbf{Blocking.} We compute $XUV$ by blocks as shown in Figure~\ref{fig:right}. The data flow is structured around a main outer loop that processes the input matrix $X$ one row strip at a time. Within each iteration of this outer loop, the computation proceeds in two phases within inner loops. First, the kernel computes and caches the entire intermediate strip, $(XU)_{m *}^T$, in SBUF. To do this, its inner loops load corresponding blocks of $X$ and $U$ from HBM. The $X$ block is transposed in transit by the DMA engine, and the blocks are multiplied, with the result $(XU)_{mp}^T$ accumulated in PSUM before being stored in the SBUF cache. In the second phase, another set of inner loops iterates through the blocks of matrix $V$, loading them from HBM and multiplying them with the pre-computed blocks fetched from the cached strip in SBUF. The final result for the output block, $O_{mn}$, is accumulated in PSUM and then written back to HBM.

The block size can impact kernel performance significantly because there is a trade-off between computational efficiency and on-chip memory usage. We model the trade-off with two metrics -- Arithmetic Intensity and Peak SBUF Usage. Assume that we have the inputs $X\in \mathbb{R}^{M\times K}$, $U\in \mathbb{R}^{K\times r}$ and $V\in \mathbb{R}^{r\times N}$ in HBM, and $s$ is the size of the data type of inputs. The arithmetic intensity (Equation ~\ref{eq:arith_inten}) is defined as the ratio of total FLOPs to HBM traffic. The HBM traffic is composed of initial reads of $X$, final writes of $O$, and repeated reads of $U$ and $V$ for each of the $M/B_M$ row strips. 
\begin{align}
\text{Arithmetic Intensity} &= \frac{2Mr(K+N)}{s \cdot \left(M(K+N) + \frac{M}{B_M}r(K+N)\right)} \nonumber \\
                           &= \frac{2r}{\left(1 + \frac{r}{B_M}\right) \cdot s} \label{eq:arith_inten}
\end{align}

The peak SBUF usage is determined by the maximum memory required across the kernel's two computational phases: first computing the intermediate strip $(XU)_{m*}$, and second, using that strip to produce the final output blocks. The peak requirement is the maximum of the SBUF footprints in these two phases, formulated as follows. 
\begin{align}
\text{Peak SBUF Usage} &= \max\left( (B_M r + B_M B_K + B_K B_r), \right. \nonumber \\
                       & \quad \left. (B_M r + B_r B_N + B_M B_N) \right) \cdot s  \nonumber \\
                       &= (B_M r  + (B_M + B_r) \cdot \max(B_K, B_N)) \cdot s \label{eq:peak_sbuf}
\end{align}

Equations 10 and 11 model a trade-off: the arithmetic intensity increases with a larger $B_M$ due to better data reuse, while the peak SBUF usage also grows as it must hold larger blocks. Therefore, an optimal block size must be large enough to maximize arithmetic intensity and saturate the tensor engine, yet small enough to fit within the 24 MB SBUF capacity to avoid memory spills. As detailed in Section~\ref{sec:block-exp}, our experimental results validate the two models.



Using Equations 10 and 11, we  determine the optimal block size. By the roofline model, a kernel becomes compute-bound when its arithmetic intensity exceeds a hardware-specific threshold \cite{williams2009roofline}—for a Trainium NeuronCore with \texttt{bfloat16} data, this threshold is 222 Flops/Byte \cite{nki_mm2025}. Setting the arithmetic intensity (Equation~\ref{eq:arith_inten}) to this threshold yields the minimum block size $B_M$ required to saturate the tensor engine. Starting with this $B_M$, the peak SBUF usage model (Equation~\ref{eq:peak_sbuf}) can then identify block combinations that maximize data reuse within the 24 MB SBUF capacity.

\subsubsection{MLP NKI Kernel}
We introduce a specialized NKI kernel for MLP layers, which extends the $XUV$ kernel design to a multi-stage operation. 

An MLP layer, such as one based on SwiGLU, consists of three matmuls.
First, two parallel linear transformations---a ``gate'' projection and an ``up'' projection---are applied to the input tensor $X$. The gate's output is passed through a SiLU activation function and then combined with the up-projection's output via an element-wise multiplication. This intermediate result is then passed through a ``down'' projection to produce the MLP layer's output.

\begin{figure}[t]
    \centering
    \includegraphics[width=0.9\linewidth]{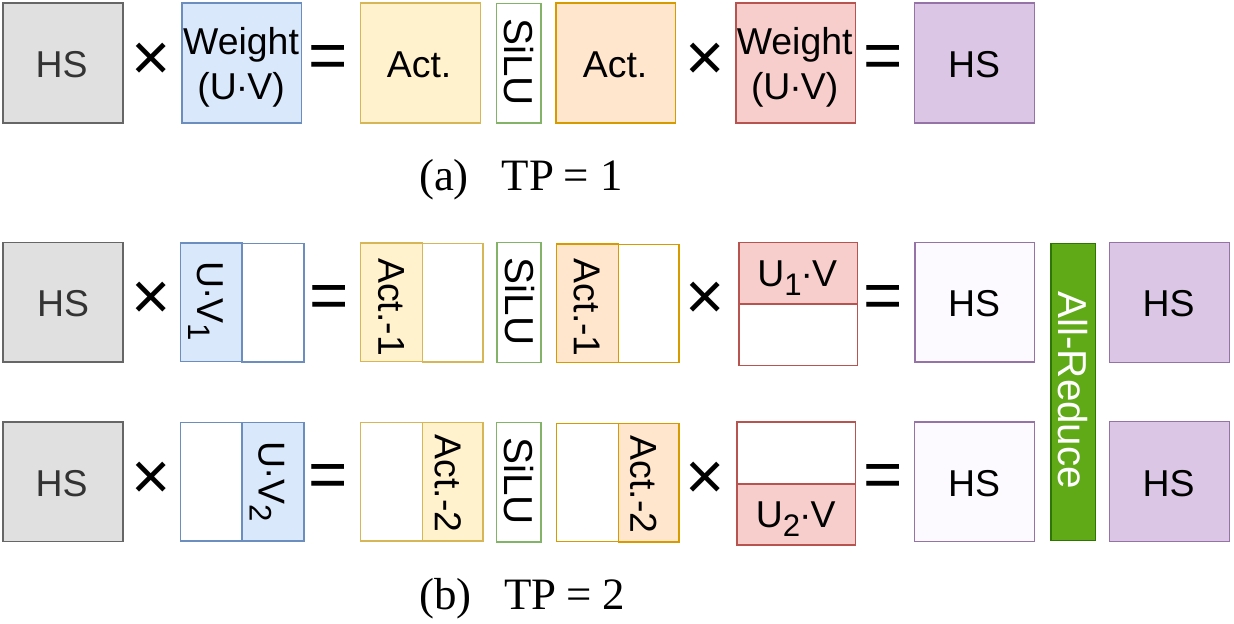}
    \caption{\name with Tensor Parallelism. V1 and V2 denote column-wise partitioning of the up projection and gate projection, while U1 and U2 denote row-wise partitioning of the down projection.}
    \label{fig:tensor_parallelism‌}
\end{figure}

Our implementation (Algorithm \ref{alg:fused_mlp_complete}) maps the MLP computation onto two NKI kernels (lines 6-7) derived from the $XUV$ kernel. The first stage, \texttt{UpGateProjection} kernel (Algorithm \ref{alg:fused_up_projection_detail}) extends the \texttt{XUV} kernel to compute the ``gate'' and ``up'' projections in parallel (lines 5-18). It then uses Trainium's Scalar and Vector Engines to perform $\texttt{SiLU}(\text{gate}) \odot \text{up}$ on-chip SBUF and writes the transposed result to HBM (line 19). The second stage, \texttt{DownProjection}, uses the $XUV$ kernel that accepts a pre-transposed input and produces a standard-layout output. This stage seamlessly consumes the transposed output from the first stage without intermediate transpose, and writes the final result to HBM in the standard layout required by subsequent LLM layers. Besides, to maximize hardware utilization of each NeuronCore, we adopt tensor parallelism~\cite{shoeybi2019megatron} as illustrated in Figure~\ref{fig:tensor_parallelism‌}. In the MLP layers, after decomposing the weight matrix into two low-rank factors, we first apply column-wise partitioning, followed by row-wise partitioning, to the decomposed weight matrices. Finally, an AllReduce operation is applied to aggregate the partial results.

\subsection{Discussions}

\subsubsection{Focus on MLP Layers in LLMs}  A typical transformer layer in LLMs consists of self-attention and MLP layers, both of which rely heavily on matrix multiplication. We apply \name to the MLP module for two primary reasons. First, MLP layers account for the majority of parameters in LLMs and are composed of higher-dimensional matrices as shown in Section~\ref{sec:data_movement}. Therefore, optimizing the MLP yields more significant reductions in both inference latency and model size. Second, when extending \name to both attention and MLP modules, we observe a substantial degradation in model accuracy, despite applying extensive fine-tuning to recover performance. As a result, we restrict the application of \name to the MLP layers only.


\subsubsection{System Lessons} In modern ML workloads, end-to-end performance is often dominated by data movement rather than raw computation, making IO-aware design an important concern. Algorithms should be structured to minimize expensive memory traffic through techniques such as tiling, on-chip computation, and kernel fusion, and co-designed with the memory hierarchy to minimize data movement between high-bandwidth memory (HBM) and on-chip SRAM. Efficient use of hardware further requires aligning computation with architectural strengths and carefully designing parallelism and work partitioning to maximize utilization and avoid data transfer overheads.

\begin{algorithm}[!tbp]
\caption{SVD-compressed MLP layer}
\label{alg:fused_mlp_complete}
\SetKwProg{Fn}{Function}{:}{} 

\footnotesize
\textbf{Require:} Matrices $X \in \mathbb{R}^{M \times K}$, $U^{\text{gate}}, U^{\text{up}}\in \mathbb{R}^{K \times r}$, $V^{\text{gate}}, V^{\text{up}} \in \mathbb{R}^{r\times N}$, $U^{\text{down}}\in \mathbb{R}^{N\times r_2}$, $V^{\text{down}}\in \mathbb{R}^{r_2\times K}$ on HBM. \textbf{Return:} Matrix $Z \in \mathbb{R}^{M \times K}$. 
\\
\textbf{Note:} Boolean \texttt{is\_XT} and \texttt{req\_OT}: input $X$ is pre-transposed, required output is $O^T$.
\\
\Fn{\texttt{XUV\_Kernel}($X, U, V, \text{is\_XT}, \text{req\_OT}$)}{
    \KwRet{$XUV$ or $(XUV)^T$}
}

\Fn{\texttt{FusedMLP}(X, $U^{\text{gate}}, U^{\text{up}},V^{\text{gate}}, V^{\text{up}}, U^{\text{down}}, V^{\text{down}}$)}{
    \BlankLine
    $Y^T \leftarrow \texttt{UpGateProjection}(X, U^{\text{gate}}, V^{\text{gate}}, U^{\text{up}}, V^{\text{up}})$\;
    $Z \leftarrow $$\texttt{XUV\_Kernel}(Y^T, U^{\text{down}}, V^{\text{down}}, \text{is\_XT=true}, \text{req\_OT=false})$;
    \KwRet{$Z$}
}
\end{algorithm}

\begin{algorithm}[!tbp]
\caption{MLP up-projection kernel}
\label{alg:fused_up_projection_detail}
\SetKwProg{Fn}{Function}{:}{}
\SetKwInput{KwInputs}{Inputs}
\SetKwInput{KwOutputs}{Outputs}
\SetKwInput{KwNote}{Note}
\footnotesize

\KwInputs{Matrices $X \in \mathbb{R}^{M \times K}$, $U^{\text{gate}}, U^{\text{up}}\in \mathbb{R}^{K \times r}$, $V^{\text{gate}}, V^{\text{up}} \in \mathbb{R}^{r\times N}$ on HBM.}
\KwOutputs{Matrix $Y^T \in \mathbb{R}^{N \times M}$.}
\KwNote{Block matrix multiplication $A\cdot B$ refers to the hardware instruction \texttt{MatmulBlock}$(s=A, m=B)$.}

\BlankLine

\Fn{\texttt{UpGateProjection}($X, U^{\text{gate}}, V^{\text{gate}}, U^{\text{up}}, V^{\text{up}}$)}{
    Allocate $Y^T \in \mathbb{R}^{N \times M}$ on HBM\;
    \For{$m \leftarrow 1$ \KwTo $\lceil M/B_M \rceil$}{
        Allocate $G_{m*},U_{m*} \in \mathbb{R}^{B_M \times r}$ on SBUF\;
        \For{$p \leftarrow 1$ \KwTo $\lceil r/B_r \rceil$}{
            Initialize $G^T_{mp},U^T_{mp} \in \mathbb{R}^{B_r\times B_M }$ on PSUM with $\mathbf{0}$\;
            \For{$k \leftarrow 1$ \KwTo $\lceil K/B_K \rceil$}{
                Load blocks $X^T_{mk}$, $U^{\text{gate}}_{kp}$ and $U^{\text{up}}_{kp}$ from HBM\; 
                \BlankLine
                $G^T_{mp} \leftarrow G_{mp} + U^{\text{gate}}_{kp} \cdot X^T_{mk}$\;
                \BlankLine
                $U^T_{mp} \leftarrow U_{mp} + U^{\text{up}}_{kp} \cdot X^T_{mk}$\;
            }
            Write $G^T_{mp}, U^T_{mp}$ to the $p$-th block of $G_{m*}, U_{m*}$\;
        }
        \For{$n \leftarrow 1$ \KwTo $\lceil N/B_N \rceil$}{
            Initialize $G^T_{mn},U^T_{mn} \in \mathbb{R}^{B_N\times B_M }$ on PSUM with $\mathbf{0}$\;
            \For{$p \leftarrow 1$ \KwTo $\lceil r/B_r \rceil$}{
                Load blocks $V^{\text{gate}}_{pn}$ and $V^{\text{up}}_{pn}$ from HBM\;
                \BlankLine
                Fetch cached block $G^T_{mp}, U^T_{mp}$ from $G_{m*},U_{m*}$\;
                \BlankLine
                $G_{mn}^T \leftarrow G_{mn}^T + V^{\text{gate}}_{pn} \cdot G^T_{mp}$\;
                \BlankLine
                $U_{mn}^T \leftarrow U_{mn}^T + V^{\text{up}}_{pn} \cdot U^T_{mp}$\;
            }
            Write $Y_{mn}^T \leftarrow \texttt{SiLU}(G_{mn}^T) \odot U_{mn}^T$ to HBM\;
        }
    }
    \KwRet{$Y^T$}
}
\end{algorithm}

\section{Evaluation}

\subsection{Experimental Setup}

\textbf{Implementation.} We develop \name on top of NeuronX Distributed Inference library~\cite{aws-neuron-neuronx-distributed-inference} and implement MLP kernels based on NKI~\cite{nki2025}. We evaluate \name on an \texttt{trn1.2xlarge} instance of Amazon Elastic Compute Cloud (EC2) equipped with AWS Trainium accelerators, running the Deep Learning Amazon Machine Images (AMI) Neuron (Ubuntu 22.04).  





\noindent\textbf{Models and datasets.} 
We test recent LLMs that fit entirely into the HBM and are currently supported by NeuronX Distributed Inference library, including Llama-3.2-1B, Llama-3.2-3B, Llama-3.1-8B, Qwen3-1.7B, Qwen3-4B, and Qwen3-8B. 
We evaluate \name with nine datasets, covering three language modeling datasets (WikiText-2~\cite{merity2016pointer}, PTB~\cite{marcinkiewicz1994building}, and C4~\cite{raffel2020exploring}) and six common sense reasoning datasets (OpenBookQA~\cite{mihaylov2018can}, WinoGrande~\cite{sakaguchi2021winogrande}, PIQA~\cite{bisk2020piqa}, HellaSwag~\cite{zellers2019hellaswag},  ARC-e, and ARC-c~\cite{clark2018think}). For fine-tuning with LoRA, we use the yahma/alpaca-cleaned~\cite{yahma_alpaca_cleaned} datasets. 


\subsection{Evaluation of $XUV$ Kernel}



We compare \name against two baselines: NKI $XW$ and NKI $XUV$. Both baselines use the state-of-the-art matmul kernel implemented by AWS official \cite{aws_nki_matrix_multiplication}. 
NKI $XW$ computes the standard matmul without SVD, while NKI $XUV$ executes matmuls using the low-rank factors $U$ and $V$ derived from the SVD of $W$, without TrainiumFusion optimization in \name. 
We evaluate our kernel with matrices $X \in \mathbb{R}^{M \times 8192}$,  $W \in \mathbb{R}^{8192 \times 16384}$, $U \in \mathbb{R}^{8192 \times 4096}$, and $V \in \mathbb{R}^{4096 \times 16384}$, where $U, V$ denotes the low-rank approximation derived from the SVD of $W$. 
We vary the first dimension $M$ of $X$ from 1024 to 32768 to simulate different sequence lengths.
For this evaluation, we assume that the kernel’s output is consumed as the stationary matrix in a subsequent computation. 
Therefore, to eliminate intermediate transpose, the kernel is configured to compute the transposed output, $O^T = (XUV)^T$.  

\begin{figure*}[!t] 
    \centering
    \includegraphics[width=\textwidth]{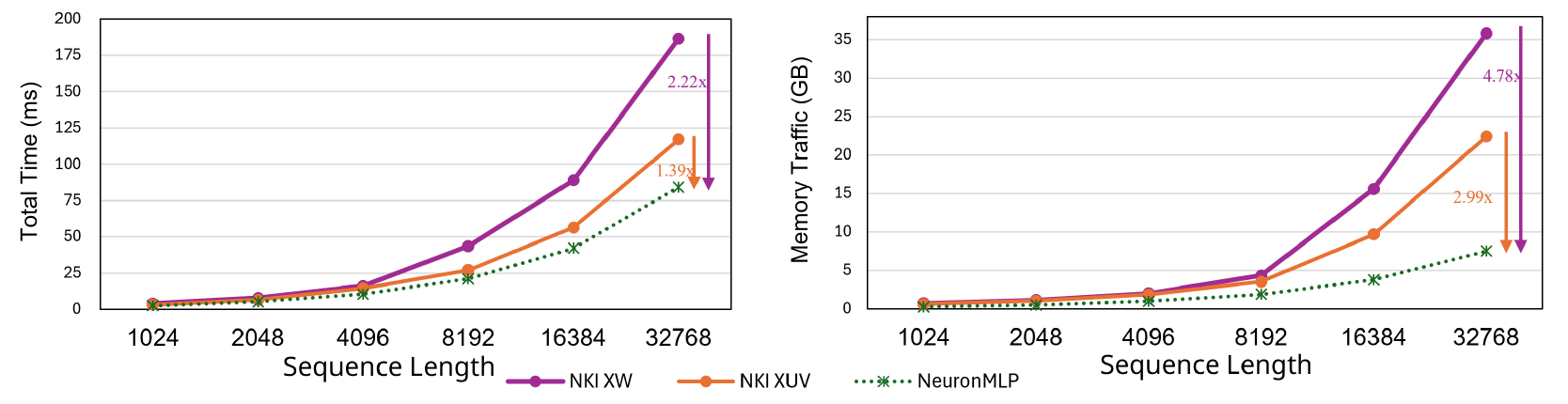} 
    \caption{Execution time and HBM-SBUF memory traffic of different matmul
implementations across input sequence lengths.}
    \label{fig:merge_t}
\end{figure*}


\begin{table}[!tb]
\centering
\renewcommand{\arraystretch}{1.3} 
\caption{Average performance across the sequence lengths from 1K to 32K. The best performance is shown in \textbf{bold}, and the second best is shown \underline{underlined}.}
\label{tab:kernel_avg}
\resizebox{1.0\linewidth}{!}{%
\begin{tabular}{l|ccc}
\toprule
 & \textbf{NKI XW} & \textbf{NKI XUV} & \textbf{\name}  \\
\midrule
Latency (ms) & 57.89 & \underline{37.47} & \textbf{27.63}  \\
Memory Traffic (GB) & 9.93  & \underline{6.52} & \textbf{2.47} \\
Tensor Engine Active Time (\%) & 78.52 & \underline{81.28} & \textbf{99.21} \\
MFU (\%) & 64.09 & \underline{65.24} & \textbf{85.20} \\
FLOPs (TFLOPS) & 2.96 & 2.18 & 2.18   \\
Transpose FLOPs (GFLOPS) & \underline{68.01} & 78.92 & \textbf{22.55} \\
\bottomrule
\end{tabular}
}
\end{table}

Figure~\ref{fig:merge_t} reports execution time and HBM-SBUF memory traffic, while Table~\ref{tab:kernel_avg} summarizes the average performance metrics for each kernel across all evaluated sequence lengths (1K--32K).
As shown in Table~\ref{tab:kernel_avg}, \name sustains the highest tensor engine active time and MFU, meaning that most cycles in the tensor engine are devoted to useful matmul operations. This high utilization directly translates into the lowest execution time, as shown in Figure~\ref{fig:merge_t}.
Compared to NKI $XW$ baseline, \name delivers an average 2.09$\times$ speedup, reaching 2.22$\times$ (84.15 ms vs. 186.60 ms) at sequence length 32K, driven by 4.78$\times$ reduction in HBM-SBUF memory traffic.  
\name also outperforms NKI $XUV$ baseline, achieving a 1.35$\times$ speedup with over 2.6$\times$ less memory traffic on average, as illustrated in Table~\ref{tab:kernel_avg}.

\textbf{Breakdown Analysis.} The performance gains of our approach stem from a software–hardware co-design that combines the algorithmic efficiency of SVD with a fused kernel tailored to Trainium. We study the contributions of SVD and TrainiumFusion separately. 

We first measure the speedup from SVD alone by comparing NKI $XW$ to NKI $XUV$. SVD yields an average speedup of 1.54$\times$. This gain results from SVD's algorithmic advantage; by factorizing $W$ into two low-rank matrices, SVD reduces both computation and memory traffic. On average, total FLOPs drop by 26\% (2.96 to 2.18 TFLOPs) and HBM–SBUF traffic decreases by 34\% (9.93 to 6.52 GB).

We then compare \name to NKI $XUV$ to isolate the effect of TrainiumFusion. \name achieves an average speedup of 1.35$\times$ by exploiting Trainium’s architecture in two ways. First, it avoids materializing the intermediate matrix in HBM, cutting the average memory traffic by 2.64$\times$ (6.52 to 2.47 GB). Second, it eliminates intermediate transposes, reducing the average transpose-related FLOPs by 3.5$\times$ (78.92 to 22.55 GFLOPs). These optimizations raise the tensor engine MFU to 85\%, compared to 65\% for the sequential kernel (i.e., NKI $XUV$).


\sloppy
Together, Block-aligned SVD’s algorithmic savings and TrainiumFusion’s hardware co-design deliver an average speedup of 2.10$\times$ over the original NKI $XW$ baseline. This demonstrates that the SVD alone is insufficient and co-designing with accelerator architecture is essential to fully realize performance gains.

\subsection{Impact of Block Size on Kernel Performance}
\label{sec:block-exp}


To empirically validate the trade-off model proposed in Section~\ref{sec:xuv-kernel}, we benchmark the \name kernel with varying block sizes, $B_M$. The experiment uses \texttt{bfloat16} matrices derived from an SVD-compressed DeepSeek-V3 MLP layer~\cite{liu2024deepseek}. We fix the input sequence length at 4096, yielding $X \in \mathbb{R}^{4096 \times 7168}$, $U \in \mathbb{R}^{7168 \times 4096}$, and $V \in \mathbb{R}^{4096 \times 18432}$.

The results in Table~\ref{tab:m_blocksize_metrics} align with our model's predictions in Section~\ref {sec:xuv-kernel}: latency decreases initially and then rises as $B_M$ grows. 
When $B_M$ increases from 128 to 1024, latency drops sharply because arithmetic intensity rises, saturating the tensor engine and making the kernel compute-bound.
For $B_M \geq 1024$, however, performance degrades as the memory footprint exceeds the SBUF capacity. This is evidenced by non-zero \texttt{spill\_reload\_bytes}. These spills stall execution and reduce arithmetic intensity.

These findings confirm that kernel performance hinges on selecting a block size that balances arithmetic intensity against on-chip memory limits. The optimal $B_M$ fully utilizes the tensor engine without triggering SBUF spills, highlighting block-size tuning as a critical design parameter for Trainium kernels.

\begin{table}[!tb]
\centering
\renewcommand{\arraystretch}{1.3} 
\caption{Performance under different block size $B_M$.}
\label{tab:m_blocksize_metrics}
\resizebox{1.0\linewidth}{!}{%
\begin{tabular}{l|cccccc}
\toprule
\textbf{$B_M$} & \textbf{128} & \textbf{256} & \textbf{512} & \textbf{1024} & \textbf{2048} & \textbf{4096} \\
\midrule
Total Time (ms) & 31.25 & 16.02 & 11.02 & 10.99 & 11.07 & 12.50 \\
Arithmetic Intensity (flops/byte) & 124.12 & 240.94 & 455.10 & 819.17 & 1280.50 & 512.95 \\
SBUF Usage (\%) & 19.54 & 51.69 & 80.07 & 90.05 & 96.35 & 98.96 \\
Spill Reload (MB) & 0 & 0 & 0 & 0 & 29.19 & 931.00 \\
Spill Save (MB) & 0 & 0 & 0 & 0 & 10.53 & 266.00 \\
\bottomrule
\end{tabular}

}
\end{table}


\subsection{Evaluation of MLP Kernel}

We integrate \name into the MLP layers of LLMs and evaluate its perplexity (PPL), accuracy, and speedup on end-to-end inference. For language modeling tasks (Wiki2, PTB, and C4), we report PPL, which measures the model’s uncertainty in predicting the next token, with lower values indicating better predictions.
For common-sense reasoning tasks (the other six tasks), we report task accuracy.

\subsubsection{Impact of Compression Ratio} 
\begin{figure}[t]
    \centering
    \includegraphics[width=1\linewidth]{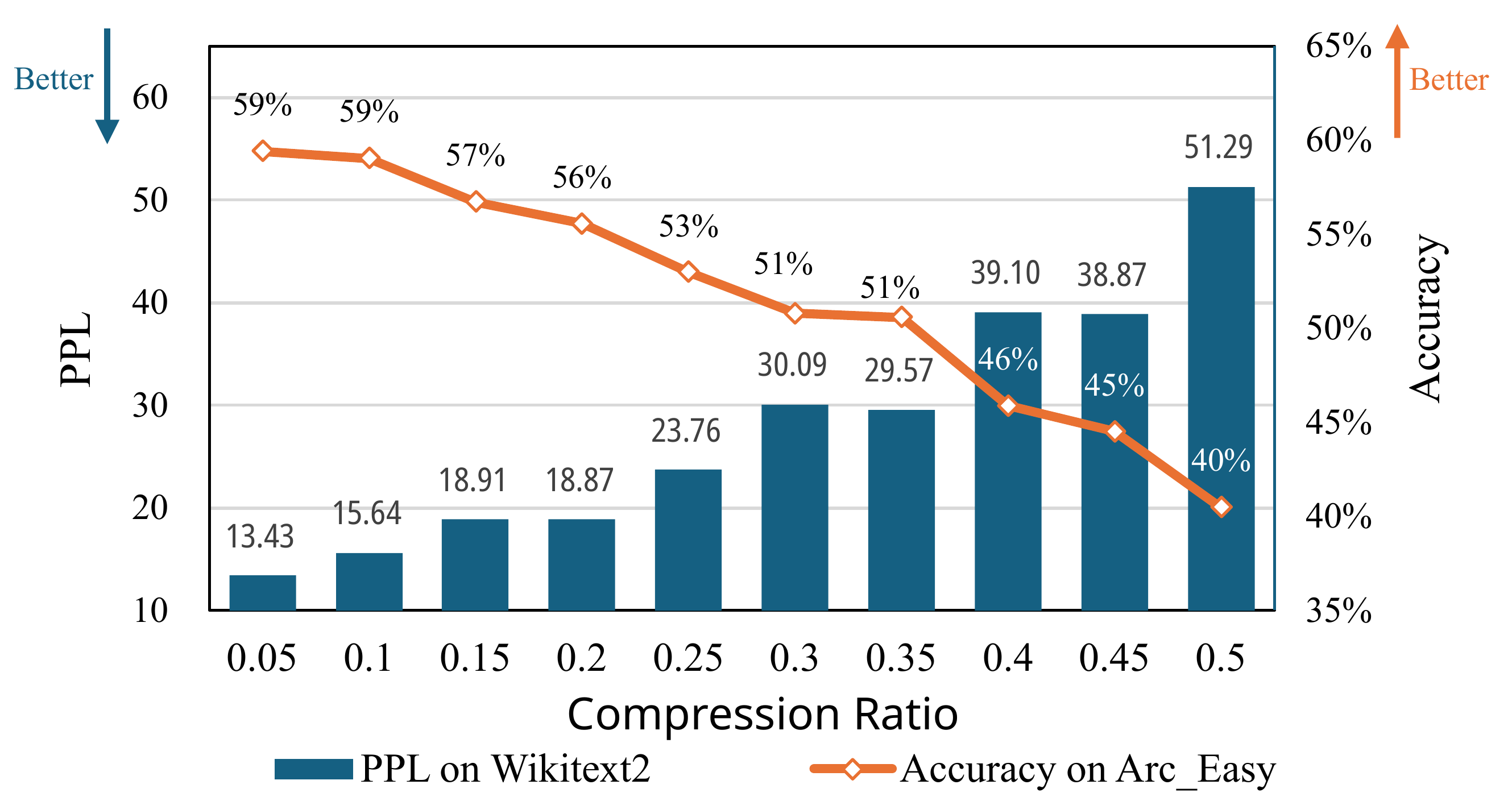}
    \caption{Model degradation with increasing compression ratios.}
    \label{fig:performance_drop}
\end{figure}

\begin{figure}[t] 
    \centering
    \includegraphics[width=\columnwidth]{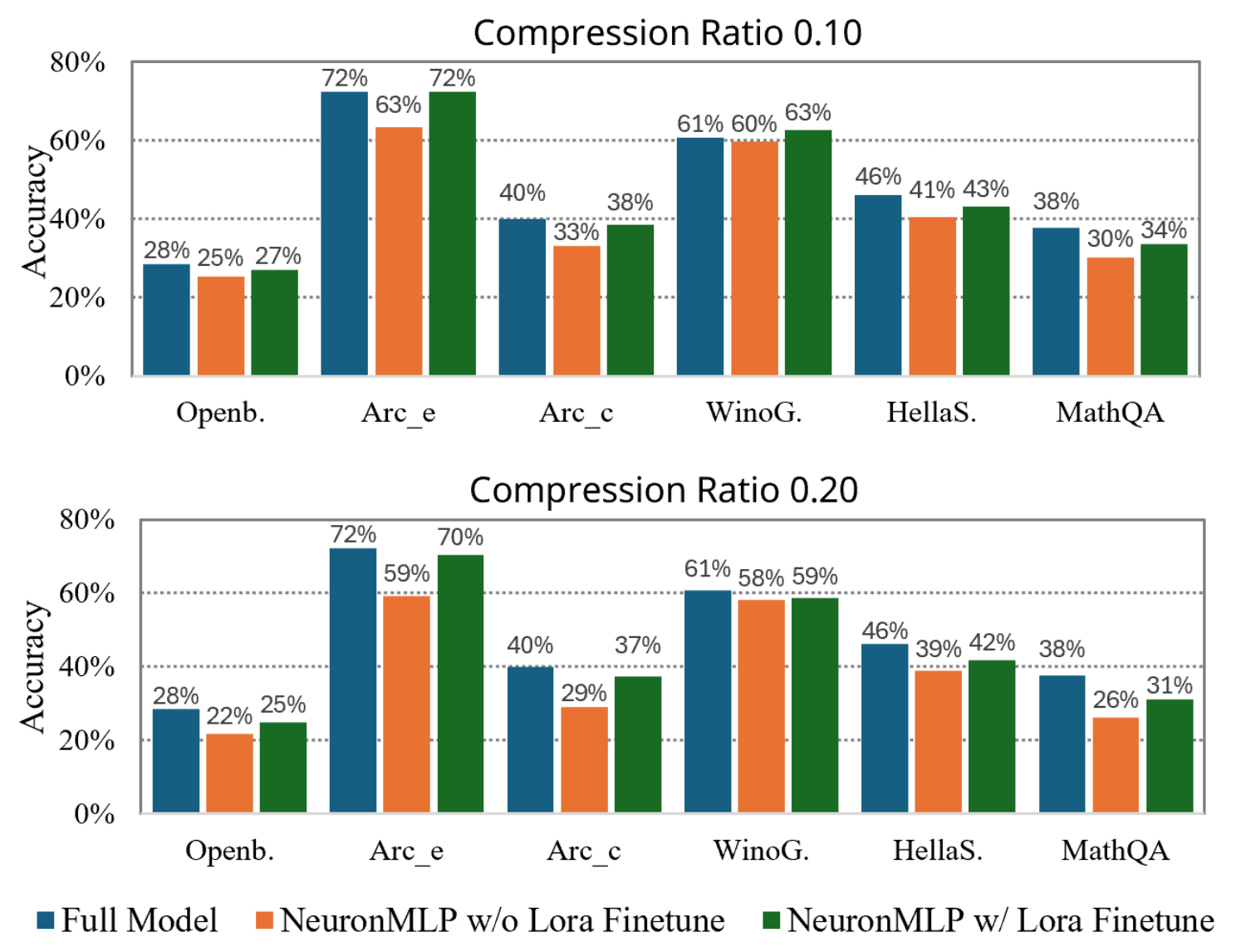} 
    \captionsetup{skip=3pt}
    \caption{Accuracy degradation and recovery of Qwen-3-1.7B under different compression ratios on six common-sense reasoning datasets.}
    \label{fig:qwen_3_1.7B_accuracy_plot}
\end{figure}

We evaluate Llama-3.2-1B under compression ratios from 0.05 to 0.5 on the WikiText-2 language modeling task and the ARC Easy commonsense reasoning benchmark. As shown in Figure~\ref{fig:performance_drop}, higher compression ratios reduce the number of model parameters but lead to increased PPL and decreased accuracy. This demonstrates the fundamental trade-off between model size and predictive performance. The model degradation is acceptable under mild compression. However, as the compression ratio increases, performance drops sharply. In particular, between 0.45 and 0.5, PPL surges from 38.87 to 51.29, and accuracy drops from 45\% to 40\%, indicating significant loss in model capability. To ensure that the accuracy degradation stays within an acceptable range~\cite{wang2024svd,wang2025svd,li2025adasvd,yuan2023asvd,wang2025dobi},  we adopt compression ratios of 0.05, 0.1, and 0.2 for the end-to-end evaluation (Section~\ref{sec:eval_llm}).  Besides, Figure~\ref{fig:qwen_3_1.7B_accuracy_plot} presents the accuracy of Qwen-3-1.7B under compression ratios of 0.1 and 0.2 across six commonsense reasoning benchmarks. We can observe that while compression reduces accuracy, LoRA fine-tuning effectively restores it, keeping degradation negligible at these low ratios.


\begin{figure}[t]
    \centering
    \includegraphics[width=1.0\linewidth]{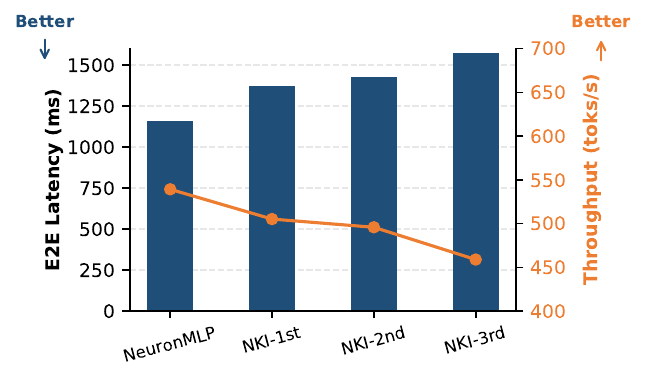}
    \caption{Comparison with NKI-based Llama-3.2-1B inference (ASPLOS/EuroSys 2025 Contest).}
    \label{fig:nki_contest}
\end{figure}

\begin{table*}[t]
\centering
\caption{Evaluation of \name across six LLMs and nine datasets.}
\setlength{\tabcolsep}{2.5pt}
\renewcommand{\arraystretch}{0.9}
\small
\begin{tabular}{c|c|ccc|cccccc|cc}
\toprule

\multirow{2}{*}{\textbf{Model}} & \multirow{2}{*}{\textbf{Compr Ratio}} & \multicolumn{3}{c|}{\textbf{PPL (↓)}} & \multicolumn{6}{c|}{\textbf{Accuracy (↑)}} & \multirow{2}{*}{\textbf{mAcc (↑)}} & \multirow{2}{*}{\textbf{Avg. Speedup (↑)}} \\
 & & Wiki2 & PTB & C4 & Openb. & ARC\_e & ARC\_c & WinoG. & HellaS. & MathQA &  &  \\
\midrule

\multirow{4}{*}{Llama-3.2-1B}& \multirow{1}{*}{Original} 
  & 9.75 & 15.40 & 13.83 & 0.26 & 0.66 & 0.31 & 0.61 & 0.48 & 0.29  & 0.43 & 1.00$\times$ \\
\cmidrule(){2-13}

& 0.05 
  & 14.32 & 17.33 & 15.46 & 0.24 & 0.63 & 0.30 & 0.59 & 0.45 & 0.29& 0.41 &1.19$\times$ \\
& 0.10 
  & 15.64 & 22.80 & 22.72 & 0.20 & 0.59 & 0.28 & 0.55 & 0.41 & 0.26& 0.38 &1.28$\times$ \\
& 0.20  
  & 18.87 &  27.24 & 26.71 & 0.18 & 0.56 &  0.26 & 0.54 & 0.39 & 0.25 & 0.36 &1.63$\times$ \\
\midrule

\multirow{4}{*}{Llama-3.2-3B} &\multirow{1}{*}{Original} 
&7.82 & 11.78 & 11.29 &0.31 & 0.74 & 0.42 &  0.70 & 0.55 & 0.35 & 0.51 & 1.00$\times$ \\
\cmidrule(){2-13}

& 0.05 
 & 9.64 & 13.52 & 14.33 & 0.29 & 0.70 & 0.42 & 0.68 & 0.51 & 0.33 & 0.49 &1.22$\times$ \\
& 0.10 
 & 11.58 & 15.61 & 17.11 & 0.24 & 0.65 & 0.34 & 0.64 & 0.47 & 0.28 & 0.44 &1.34$\times$ \\
& 0.20  
  & 15.13 & 18.69 & 20.80 & 0.23 & 0.62 & 0.29 & 0.60 & 0.43 & 0.27 & 0.41 &1.78$\times$ \\
  
\midrule

\multirow{4}{*}{Llama-3.1-8B} &\multirow{1}{*}{Original} 
&6.99 & 10.75 & 10.94 &0.43 & 0.80 & 0.51 &  0.73 & 0.59 & 0.38 & 0.57 & 1.00$\times$ \\
\cmidrule(){2-13}

& 0.05 
 & 8.56 & 11.43 & 13.78 & 0.40 & 0.76 & 0.47 & 0.72 & 0.54 & 0.36 & 0.54 &1.24$\times$ \\
& 0.10 
 & 11.03 & 13.59 & 18.32 & 0.34 & 0.69 & 0.42 & 0.67 & 0.49 & 0.33 & 0.49 &1.37$\times$ \\
& 0.20  
  & 14.73 & 16.25 & 22.78 & 0.29 & 0.66 & 0.37 & 0.62 & 0.45 & 0.30 & 0.45 &1.69$\times$ \\
  
\midrule

\multirow{3}{*}{Qwen-3-1.7B} & \multirow{1}{*}{Original} 
 & 16.68 & 28.88 &  22.80 & 0.28 & 0.72 & 0.40 & 0.61 & 0.46 & 0.38 & 0.47 & 1.00$\times$ \\
\cmidrule(){2-13}
& 0.05
  & 16.54 & 26.33 & 21.12 & 0.28 & 0.72 & 0.38 & 0.60 & 0.45 & 0.37 & 0.47 & 1.29$\times$ \\
& 0.10 
  & 15.43 & 25.00 & 23.16 & 0.27 & 0.72 & 0.39 & 0.62 & 0.43 & 0.34 & 0.46 & 1.41$\times$ \\
& 0.20 
 & 17.05 & 26.97 & 25.14 & 0.25 & 0.70 & 0.37 & 0.58 & 0.42 & 0.31 & 0.44 & 1.74$\times$ \\
\midrule

\multirow{3}{*}{Qwen-3-4B} & \multirow{1}{*}{Original} 
 & 9.75 & 15.40 &  13.83 & 0.29 & 0.80 & 0.51 & 0.66 &  0.52 & 0.47  & 0.54 & 1.00$\times$ \\
\cmidrule(){2-13}

&0.05 
 & 11.23 & 16.32 & 15.45 & 0.31 & 0.79 & 0.49 & 0.68 & 0.52 & 0.44 & 0.53 & 1.17$\times$ \\

&0.10 
 & 12.18 & 18.98 & 19.05 & 0.31 & 0.78 & 0.47 & 0.66 & 0.50 & 0.42 & 0.52 & 1.25$\times$ \\
&0.20 
 & 14.05 & 21.09 & 21.38 & 0.30 & 0.75 & 0.43 & 0.64 & 0.48 & 0.37 & 0.49 & 1.67$\times$ \\

\midrule

\multirow{3}{*}{Qwen-3-8B} & \multirow{1}{*}{Original} 
 & 8.02 & 12.25 &  11.92 & 0.42 & 0.82 & 0.54 & 0.69 &  0.56 & 0.45  & 0.58 & 1.00$\times$ \\
\cmidrule(){2-13}

&0.05 
 & 10.56 & 14.12 & 13.67 & 0.37 & 0.79 & 0.52 & 0.69 & 0.53 & 0.45 & 0.56 & 1.18$\times$ \\

&0.10 
 & 13.25 & 16.23 & 16.23 & 0.35 & 0.80 & 0.51 & 0.68 & 0.52 & 0.44 & 0.55 & 1.28$\times$ \\
&0.20 
 & 15.67 & 19.07 & 20.89 & 0.31 & 0.78 & 0.47 & 0.62 & 0.50 & 0.40 & 0.51 & 1.65$\times$ \\

\bottomrule
\end{tabular}

\label{tab:overall_performance}
\end{table*}

\subsubsection{Impact of Different LLMs}
\label{sec:eval_llm}
To demonstrate the generalization ability of \name across different LLMs, we evaluate it on six LLMs and nine datasets with a tensor parallelism degree of 2. We report mean accuracy (mAcc) and average end-to-end speedup, with results summarized in Table~\ref{tab:overall_performance}. 
Across all evaluated LLMs, the SVD-compressed models retain accuracy largely comparable to the original, with mAcc drop $\leq$ 0.10 in every case, a level of loss generally considered acceptable~\cite{wang2024svd,wang2025svd,li2025adasvd,yuan2023asvd,wang2025dobi}. At a compression ratio of 0.05, \name achieves an average 1.21$\times$ end-to-end inference speedup across six LLMs. As the compression ratio increases from 0.05 to 0.20, NeuronMM delivers substantial end-to-end performance gains, with speedups ranging from 1.17$\times$ to 1.78$\times$. For instance, on Qwen-3-1.7B, \name achieves 1.74$\times$ faster inference while incurring only a 0.03 average accuracy drop compared to standard LLM inference.

\subsubsection{Comparison with the NKI Llama Inference}

We compare \name with the latest NKI-based Llama-3.2-1B inference system from the ASPLOS/EuroSys 2025 Contest~\cite{aws-neuron-nki-llama}, evaluating against the top three submissions. These implementations apply optimizations such as GEMM/GEMV tiling, instruction fusion, and kernel fusion. Figure~\ref{fig:nki_contest} shows the comparison results. We use a sequence length of 640 and a tensor parallelism degree of 2, following the contest setting, and set the compression ratio to 0.05. As shown in Figure~\ref{fig:nki_contest}, \name achieves 15.6\%, 18.8\%, and 26.3\% lower latency, and 6.7\%, 8.8\%, and 17.5\% higher throughput compared to the 1st-, 2nd-, and 3rd-place baselines, respectively.



\section{Related Work}

\textbf{SVD Compression for LLMs.}
LLMs can have billions of parameters, making inference on resource-constrained hardware challenging. Various model compression techniques have been proposed to reduce their latency and memory footprint~\cite{ dettmers2023qloraefficientfinetuningquantized,Frantar2023OPTQAQ, wu2023understandingint4quantizationtransformer, frantar2023sparsegpt, ma2023llmprunerstructuralpruninglarge, guo2023compressostructuredpruningcollaborative,ashkboos2024slicegpt, DynaBERTHou,miles2022informationtheoreticrepresentationdistillation,yang2022understandingprivilegedfeaturesdistillation}. Recent works have explored various SVD-based approaches for model compression. For example, FWSVD~\cite{hsu2022language} introduces a weighted low-rank factorization scheme, while ASVD~\cite{yuan2023asvd} develops an activation-aware SVD technique that exploits layer-wise activation patterns to enhance compression effectiveness. SVD-LLM~\cite{wang2024svd, wang2025svd} further incorporates truncation-aware data whitening and layer-specific parameter updates, assigning unique ratios to individual weights. In contrast, Dobi-SVD~\cite{wang2025dobi} establishes a bijection mapping, allowing the model to automatically learn the best truncation points among layers. However, those works do not consider the characteristics of accelerator hardware and the additional matmul overhead introduced by SVD (e.g., memory spilling) as \name.

\textbf{Performance optimization on Trainium.} Recent studies~\cite{xue2024ninjallmfastscalablecosteffective, biran2025adaptiveorchestrationlargescaleinference} have demonstrated that AWS Trainium is promising to accelerate generative AI workloads. HLAT \cite{fan2024hlat}  demonstrates that, through a series of system-level optimizations and training strategies, Trainium can reduce training costs to 60\% of those on p4d GPU instances while maintaining comparable model quality to GPU-based baselines. Complementing this, a study~\cite{fu2024distributed} details NeuronX Distributed Training (NxDT), quantifies scaling and efficiency against contemporary GPU baselines, and elucidates runtime and compiler support critical for stable large-cluster operation. However, existing work primarily focuses on pretraining, with limited exploration of inference on Trainium.

\section{Conclusions}

We present NeuronMLP, an efficient LLM inference approach on Trainium. \name is a combination of SVD compression and architecture-specific optimization for high performance.  Our evaluation demonstrates that, at a compression ratio of 0.05, NeuronMM consistently accelerates LLM inference across nine datasets and six recent LLMs, achieving an average 1.35$\times$ speedup at the matmul kernel level and an average 1.21$\times$ speedup for end-to-end LLM inference.




\balance
\bibliographystyle{ACM-Reference-Format}
\bibliography{bib/sherry, bib/li, bib/su,bib/dinghong}
\end{document}